\documentclass[11pt]{article}
\usepackage{eacl2017}
\usepackage{times}
\usepackage{url}
\usepackage{latexsym}
\usepackage{graphicx}
\usepackage{subfig}
\usepackage{amsmath}

\eaclfinalcopy 

\newcommand{\ignore}[1]{}

\title{The BURCHAK corpus: a Challenge Data Set for Interactive Learning of Visually Grounded Word Meanings}

\author{Yanchao Yu\\ 
  Interaction Lab   \\
 Heriot-Watt University  \\
  {\tt y.yu@hw.ac.uk} \\ \And
  Arash Eshghi \\
 Interaction Lab   \\
 Heriot-Watt University  \\
  {\tt a.eshghi@hw.ac.uk} \\ \And
  Gregory Mills \\
 University of Groningen  \\
  {\tt g.j.mills@rug.nl} \\ \And
  Oliver Lemon \\
 Interaction Lab   \\
 Heriot-Watt University  \\
  {\tt o.lemon@hw.ac.uk}}

\date{}

\begin{document}
\maketitle
\begin{abstract}
We motivate and describe a new freely available  human-human dialogue data set for interactive learning of visually grounded word meanings through ostensive 
definition by a tutor to a learner. The data has been collected using a novel, character-by-character variant of \textit{the DiET chat tool} \cite{Healey.etal03,Mills.HealeySubmitted} 
with a novel task, where a Learner needs to learn invented visual attribute words (such as ``burchak'' for square) from a tutor. As such, the text-based interactions closely resemble face-to-face conversation and thus contain many of the linguistic phenomena encountered in natural, spontaneous dialogue. These include self- and other-correction, mid-sentence continuations, interruptions,  overlaps, fillers, and hedges. We also present a generic n-gram framework for building user (i.e.\ tutor) simulations from this type of incremental  data, which is freely available to researchers. We show that the simulations produce outputs that are similar to the original data (e.g.\ 78\% turn match similarity). Finally, we train and evaluate a Reinforcement Learning dialogue control agent for learning visually grounded word meanings, trained from the BURCHAK corpus. The learned policy shows comparable performance to a rule-based system built previously.


\end{abstract}

\section{Introduction}

Identifying, classifying, and talking about objects and events in the surrounding environment are key capabilities for intelligent, goal-driven systems that interact with other humans and the external world (e.g.\  robots,  smart spaces, and other automated systems). To this end, there has recently been a surge of interest and significant progress made on a variety of related tasks, including generation of Natural Language (NL) descriptions of images, or identifying images based on NL descriptions  \cite{bruni2014multimodal,socher2014grounded,Farhadi09describingobjects,silberer-lapata:2014:P14-1,sun2013attribute}. Another strand of work has focused on incremental reference resolution in a model where word meaning is modeled as classifiers (the so-called Words-As-Classifiers model \cite{Kennington.Schlangen15}).

\begin{figure}\centering
\centering
\begin{tabular}{l}
\\
\begin{tabular}[c]{@{}l@{}}T(utor): it is a ... [[sako]] burchak.\\
                           L(earner): \qquad\hspace{0.1cm}   [[suzuli?]]\\          
                           T: no, it's sako \\ 
                           L: okay, i see. 
\end{tabular} \\   
\multicolumn{1}{c}{\small (a) Dialogue Example from the corpus} \\
\hline
\begin{tabular}[c]{@{}l@{}}\centering\raisebox{-1cm}{\includegraphics[width=0.95\linewidth]{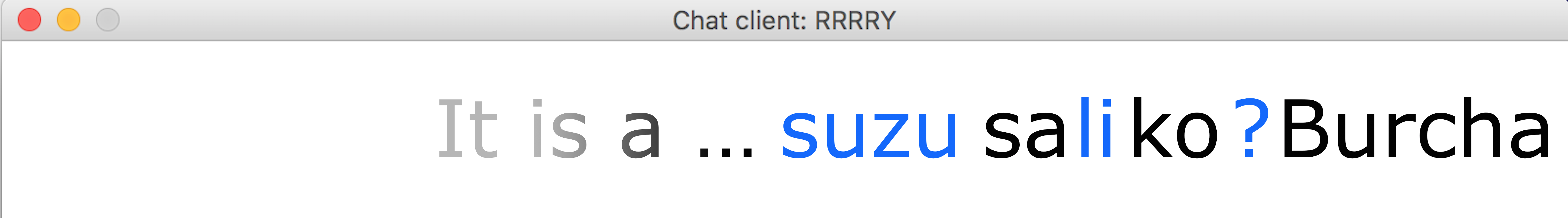}}
\end{tabular} \\\\
\multicolumn{1}{c}{\small (b) The Chat Tool Window during dialogue in (a) above}

\end{tabular}
\vspace{-0.3cm}
\caption{Example of turn overlap + subsequent correction in the BURCHAK corpus (`sako' is the invented word for red, `suzuli' for green and `burchak' for square)}
\label{fig:corpus_example}
\vspace{-0.2cm}
\end{figure}

However, none of this prior work focuses on how concepts/word meanings are {\it learned and adapted in interactive dialogue} with a human, the most common setting in which robots, home automation devices, smart spaces etc.\ operate, and, indeed the richest resource that such devices could exploit for adaptation over time to the idiosyncrasies of the language used by their users.

Though recent prior work has focused on the problem of learning visual groundings in interaction with a tutor (see e.g. \cite{yu-eshghi-lemon:2016:sigdial,Yu.etal16inlg}), it has made use of hand-constructed, synthetic dialogue examples that thus lack in variation, and many of the characteristic, but consequential phenomena observed in naturalistic dialogue (see below). Indeed, to our knowledge, there is no existing data set of real human-human dialogues in this domain, suitable for training multi-modal conversational agents that perform the task of \emph{actively learning visual concepts} from a human partner in \emph{natural, spontaneous} dialogue.

\begin{table}[!ht]
\centering
\begin{tabular}{l}
\hline
\multicolumn{1}{c}{\textbf{(a) Multiple Dialogue Actions in one turn}} \\ \hline\hline
\begin{tabular}[c]{@{}l@{}}L: so this shape is wakaki?\\ 
						   T: yes, well done. let's move to the color. \\              
                           \quad So what color is this?
\end{tabular} \\ \hline

\multicolumn{1}{c}{\textbf{(b) Self-Correction}} \\ \hline\hline
\begin{tabular}[c]{@{}l@{}}L: what is this object?\\ 
                           T: this is a sako ... no no ... a suzuli burchak. 
\end{tabular} \\ \hline

\multicolumn{1}{c}{\textbf{(c) Overlapping}} \\ \hline\hline
\begin{tabular}[c]{@{}l@{}}T: this color [[is]] ... [[sa]]ko.\\
                           L:\qquad \qquad [[su]]zul[[i?]] \\          
                           T: no, it's sako. \\ 
                           L: okay. 
\end{tabular} \\ \hline                          

\multicolumn{1}{c}{\textbf{(d) Continuation}} \\ \hline\hline
\begin{tabular}[c]{@{}l@{}}T: what is it called? \\
                           L: sako \\      
                           T: and? \\
                           L: aylana. 
\end{tabular} \\ \hline       

\multicolumn{1}{c}{\textbf{(e) Fillers}} \\ \hline\hline
\begin{tabular}[c]{@{}l@{}}T: what is this object?\\
                           L: a sako um... sako wakaki. \\   
                           T: great job.  
\end{tabular} \\ \hline              

\end{tabular}
\caption{Dialogue Examples in the Data (L for the learner and T for the tutor)\label{tab:examples}}
\end{table}

Natural, spontaneous dialogue is \emph{inherently incremental} \cite{Crocker.etal00,Ferreira96,Purver.etal09}, and thus gives rise to dialogue phenomena such as self- and other-corrections, continuations, unfinished sentences, interruptions and overlaps, hedges, pauses and fillers. These phenomena are interactionally and semantically consequential, and contribute directly to how dialogue partners coordinate their actions and the emergent semantic content of their conversation. They also strongly mediate how a conversational agent might adapt to their partner over time. For example, self-interruption, and subsequent self-correction (see example in table\ \ref{tab:examples}.b) as well as hesitations/fillers (see example in table\ \ref{tab:examples}.e) aren't simply noise and are used by listeners to guide linguistic processing \cite{Clark.FoxTree02}; similarly, while simultaneous speech is the bane of dialogue system designers, interruptions and subsequent continuations (see examples in table\ \ref{tab:examples}.c and \ref{tab:examples}.d) are performed deliberately by speakers to demonstrate strong levels of understanding \cite{Clark96}.

Despite this importance, these phenomena are excluded in many dialogue corpora, and glossed over/removed by state of the art speech recognisers (e.g.\ Sphinx-4 \cite{walker:2004:SFO} and Google's web-based ASR \cite{schalkwykBeefermanEtAl:2010:springer}; see \newcite{Baumann.etal16} for a comparison). One reason for this is that naturalistic spoken interaction is excessively expensive and time-consuming to transcribe and annotate on a level of granularity fine-grained enough to reflect the strict time-linear nature of these phenomena.

In this paper, we present a new dialogue data set - the BURCHAK corpus - collected using a new \emph{incremental} variant of the DiET chat-tool \cite{Healey.etal03,Mills.HealeySubmitted}\footnote{Available from \url{https://sites.google.com/site/hwinteractionlab/babble}}, which enables character-by-character, text-based interaction between pairs of participants, and which circumvents all transcription effort as all this data, including all timing information at the character level is automatically recorded.

\begin{figure}[!th]
\centering
\subfloat[Chat Client Window (Tutor: ``it is a ...", Learner: ``suzuli?",   Tutor: ``sako burch" ) \label{fig:chatWindow}]
  {\includegraphics[width=\linewidth]{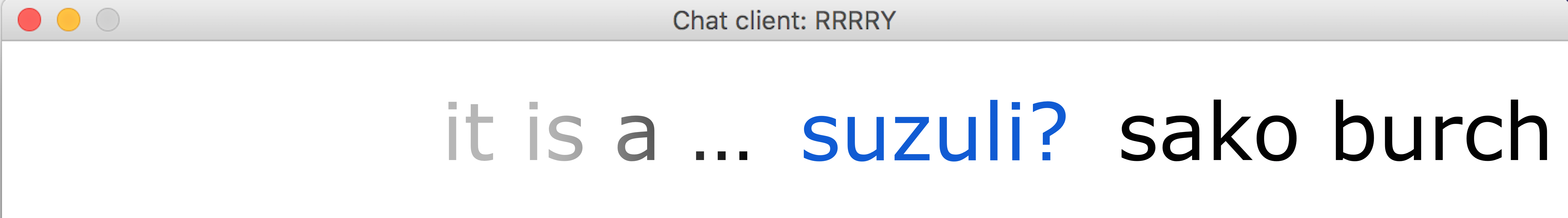}}\hfill\vspace{0.3cm}
\subfloat[Task Panel for Tutor (Learner only sees the object)\label{fig:taskPanel}]
  {\includegraphics[width=0.95\linewidth]{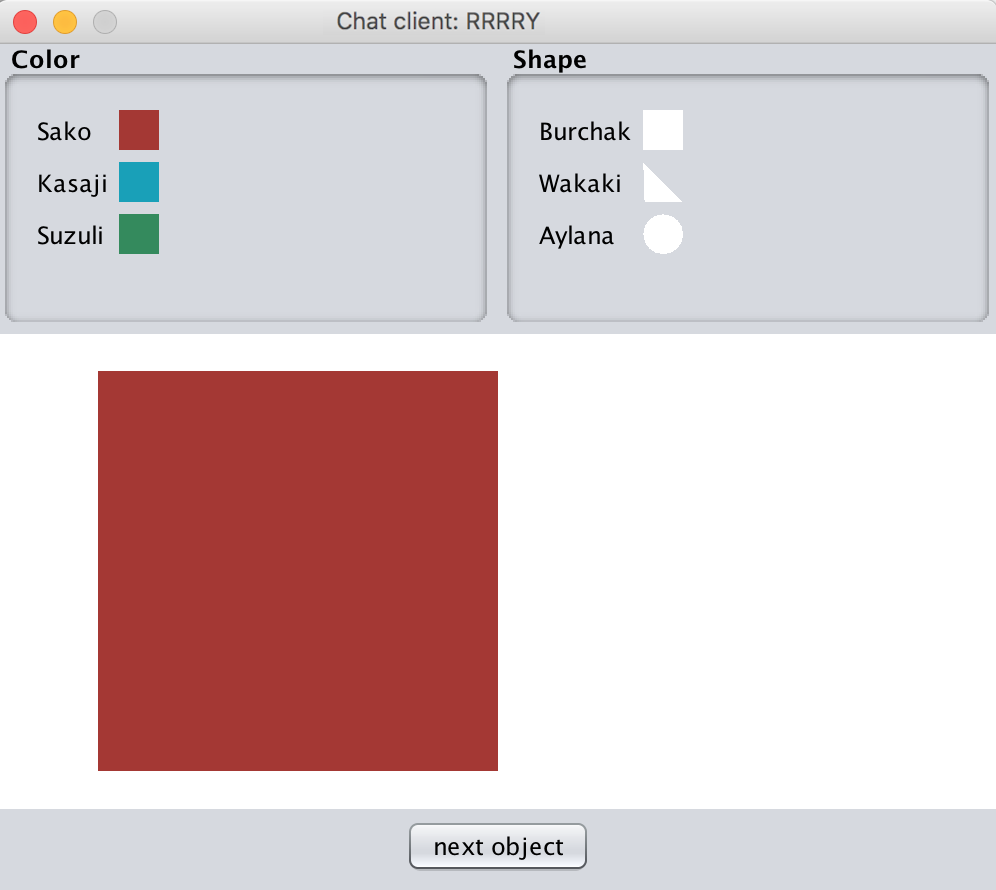}}\hfill
\caption{Snapshot of the DiET Chat tool, the Tutor's Interface \label{fig:chattool}}
\end{figure}

The chat-tool is designed to support, elicit, and record at a fine-grained level, dialogues that resemble the face-to-face setting in that turns are: (1) constructed and displayed incrementally as they are typed; (2) transient; (3) potentially overlapping as participants can type at the same time; (4) not editable, i.e.\ deletion is not permitted - see Sec. \ref{sec:chattool} and Fig.\ \ref{fig:chattool}. Thus, we have been able to collect many of the important phenomena mentioned above that arise from the inherently incremental nature of language processing in dialogue - see table \ref{fig:corpus_example}.


Having presented the data set, we then go on to introduce a generic n-gram framework for building user simulations for either task-oriented or non-task-oriented dialogue systems from this data-set, or others constructed using the same tool. 
We apply this framework to train a robust user model that is able to simulate the tutor's behaviour to interactively teach (visual) word meanings to a Reinforcement Learning dialogue agent.

\vspace{-0.2cm}
\section{Related Work}
In this section, we will present an overview of relevant data-sets and  techniques for Human-Human dialogue collection, as well as approaches to user simulation based on   realistic data. 

\subsection{Human-Human Data Collection}

There are several existing corpora of human-human spontaneous spoken dialogue, such as SWITCHBOARD \cite{Godfrey.etal92}, and the British National Corpus, which consist of open, unrestricted telephone conversations between people, where there are no specific tasks to be achieved. These datasets contain many of the incremental dialogue phenomena that we are interested in, but there is  no shared visual scene between participants, meaning we cannot use such data to explore learning of perceptually grounded language. More relevant is the MAPTASK corpus \cite{maptask}, where dialogue participants both have maps which are not shared. This dataset allows investigation of negotiation dialogue, where object names can be agreed, and so does support some work on language grounding. However, in the MAPTASK, grounded word meanings are not taught by ostensive definition as is the case in our new dataset.

We further note that the DiET Chat Tool \cite{Healey.etal03,Mills.HealeySubmitted}
 while designed to elicit conversational structures which resemble face-to-face dialogue (see examples in table\ \ref{tab:examples}), circumvents the need for the very expensive and time-consuming step of spoken dialogue transcription, but nevertheless produces data at a very fine-grained level. It also includes tools for creating more abstract (e.g. turn-based) representations of conversation.

\vspace{-0.4cm}
\subsection{User Simulation}
\vspace{-0.3cm}

Training  a dialogue strategy is one of the fundamental tasks of the user simulation. Approaches to user simulation can be categorised based on the level of abstraction at which the dialogue is modeled: 1) the intention-level has become the most popular user model that predicts the next possible user dialogue action according to the dialogue history and the user/task goal \cite{eckert:1997:ieee,asri:16:corr,cuayahuitl:2005:human,chandramohan:2012:behavior,eshkyAS:2012:emnlp,aiW:2008:sigdial,georgilaHL:2005:interspeech}; 2) on the word/utterance-level, instead of dialogue action, the user simulation can also be built for predicting the full user utterances or a sequence of words given specific information \cite{chung:2004:acl,schatzmannY:2007:asru}; and 3) on the semantic-level, the whole dialogue can be modeled as a sequence of user behaviors in the semantic representation \cite{schatzmannTWYY:2007:naacl,schatzmannY:2007:sigdial,Kalatzis.etal16}.

There are also some user simulations built on multiple levels. For instance, \newcite{jungLKJL:2009:csl} integrated different data-driven approaches on intention and word levels to build a novel user simulation. The user intent simulation is for generating user intention patterns, and then a two-phase data-driven domain-specific user utterance simulation is proposed to produce a set of structured utterances with sequences of words given a user intent and select the best one using the BLEU score. 
The user simulation framework we present below is generic in that one can use it to train user simulations on a word-by-word, utterance-by-utterance, or action-by-action levels, and it can be used for both goal-oriented and non-goal-oriented domains. 

\section{Data Collection using the DiET Chat Tool and a Novel Shape and Colour Learning Task \label{sec:dataCollection}}
\label{sec:chattool}
In this section, we describe our data collection method and process, including the concept learning task given to the human participants. 
\vspace{-0.1cm}
\paragraph{The DiET experimental toolkit} This is a custom-built Java application \cite{Healey.etal03,Mills.HealeySubmitted} 
that allows two or more participants to communicate in a shared chat window. It supports live, fine-grained and highly local experimental manipulations of ongoing human-human conversation (see e.g. \cite{Eshghi.Healey15}). The variant we use here supports text-based, character-by-character, interaction between pairs of participants, and here we use it solely for data-collection, where everything that the participants type to each other passes through the DiET server, which transmits the utterance to the other clients on the character level and all are displayed \emph{on the same row/track} in the chat window (see Fig. \ref{fig:chatWindow}) - this means that when participants type at the same time in interruptions and turn overlaps, their utterances will be all jumbled up (see Fig.\ \ref{fig:corpus_example}b). To simulate the transience of speech in face-to-face conversation with its characteristic phenomena, all utterances in the chat window   fade out after 1 second.
 Furthermore, like in speech, deletes are not permitted: if a character is typed, it cannot be deleted. The chat-tool is thus designed to support, elicit, and record at a fine-grained level, dialogues that resemble face-to-face dialogue in that turns are: (1) constructed and displayed incrementally as they are typed; (2) transient; (3) potentially overlapping; (4) not editable, i.e.\ deletion is not permitted.

\vspace{-0.1cm}
\paragraph{Task and materials} 
The learning/tutoring task given to the participants involves a pair of participants who talk about visual attributes (e.g.\ colour and shape) through a sequence of 9 visual objects, one at a time. The objects are created based on a 3 x 3 visual attribute matrix (including 3 colours and 3 shapes (see Fig.\ref{fig:taskPanel})). This task is assumed in a second-language learning scenario, where each visual attribute, instead of standard English words, is assigned to a new unknown word in a made-up language, e.g.\ ``sako'' for red and ``burchak'' for square: participants are not allowed to use any of the usual colour and shape words from the English language. 
We design the task in this way to collect data for situations where a robot has to learn the meaning of human visual attribute terms. In such a setting the robot has to learn the perceptual groundings of words such as ``red". However, humans already know these groundings, so to collect data about teaching such perceptual meanings, we invented new attribute terms whose groundings the Learner must discover through interaction. 

The overall goal of the task is for the learner to identify the shape and colour of the presented objects correctly for as many objects as possible. So the tutor initially needs to teach the learner about these using the presented objects. For this, the tutor is provided with a visual dictionary of the (invented) colour and shape terms (see Fig.\ \ref{fig:chattool}), but the learner only ever sees the object itself. The learner will thus gradually learn these and be able to identify them, so that initiative in  the conversation tends to be reversed on later objects, with the learner making guesses and the tutor either confirming these or correcting them.

\vspace{-0.1cm}
\paragraph{Participants}
Forty participants were recruited from among students and research staff from various disciplines at Heriot-Watt 
 University, including 22 native speakers and 18 non-native speakers.

\vspace{-0.1cm}
\paragraph{Procedure}
The participants in each pair were randomly assigned to experimental roles (Tutor vs.\ Learner). They were given written instructions about the task  and had an opportunity to ask questions about the procedure. They were then seated back-to-back in the same room, each at a desk with a PC displaying the appropriate task window and chat client window (see Fig.\ref{fig:chattool}). They were asked to go through all visual objects in at most 30 minutes and then the Learner was assessed to check how many new colour and shape words they had learned. Each participant was paid £$10.00$ for participation. The best performing pair was also given a £20 Amazon Voucher as prize.

\section{The BURCHAK Corpus Statistics\label{sec:statistic}}
\subsection{Overview\label{sec:overview}}
Using the above procedure, we have collected 177 dialogues (each about one visual object) with a total of 2454 turns, where a turn is defined\footnote{Note that the definition of a `turn' in an incremental system is somewhat arbitrary.} as a sequence of consecutive characters typed by a single participant with a delay of no more than 1100 ms 
 between the characters. Figure \ref{fig:dialogLength} shows the distribution of dialogue length (i.e.\ number of turns) in the corpus. where the average number of turns per dialogue is $13.86$.

\subsection{Incremental Dialogue Phenomena\label{sec:incremental}}

As noted, the DiET Chattool is designed to elicit and record conversations that resemble face-to-face dialogue. In this paper, we report specifically on a variety of dialogue phenomena that arise from the incremental nature of language processing. These are the following:

\begin{itemize}
  \vspace{-0.2cm}
  \item \textbf{Overlapping}: where interlocutors speak/type at the same time (i.e. the original corpus contains over 800 overlaps), leading to jumbled up text on the DiET interface (see Fig.\ \ref{fig:corpus_example});
  
  \vspace{-0.2cm}
  \item  \textbf{Self-Correction}: a kind of correction that is performed incrementally in the same turn by a speaker; this can either be conceptual, or simply repairing a misspelling or mis-pronunciation.

  \vspace{-0.2cm}
  \item \textbf{Self-Repetition}: the interlocutor repeats words, phrases, even sentences, in the same turn.
  
  \vspace{-0.2cm}
  \item \textbf{Continuation (aka Split-Utterance)}: the interlocutor continues the previous utterance (by herself or the other) where either the second part, or the first part or both are syntactically incomplete.

  \vspace{-0.2cm}
  \item \textbf{Filler}: allows the interlocutor to further plan her  utterance while keeping the floor. These can also elicit continuations from the other \cite{Howes.etal12b}. This is performed using tokens such as `urm', `err', `uhh', or `\ldots'.
\end{itemize}

For annotating self-corrections, self-repetitions and continuations we have loosely followed protocols from \newcite{Purver.etal09,Colman.Healey11}. Figure \ref{fig:phenomenaFrequency} shows how frequently these incremental phenomena occur in the BURCHAK Corpus. This figure excludes Overlaps which were much more frequent: $800$ in total, which amounts to about $4.5$ per dialogue.

\subsection{Cleaning up the data for the User Simulation}

For the purpose of the annotation of Dialogue Actions, subsequent training of the user simulation, and the Reinforcement Learning  described below, we cleaned up the original corpus as follows: 1) we fixed the spelling mistakes which were not repaired by the participants themselves; 2) we also removed snippets of conversation where the participants had misunderstood the task (e.g. trying to describe the objects or where they had used other languages) (see Figure \ref{fig:misunderstood}); as well as 3) removing emoticons (which frequently occurs in the chat tool). 

\begin{figure}[h]
\begin{tabular}{|l|}
\hline
\begin{tabular}[|c|]{@{}l@{}}
						   T: the word for the color is similar to the word\\ 
							  \quad for Japanese rice wine. except it ends in o.\\
                           L: sake? \\          
                           T: yup, but end with an o. \\
                           L: okay, sako.
\end{tabular} \\ \hline 
\end{tabular}
\vspace{-0.3cm}
\caption{Example of Dialogue Snippet with the misunderstanding of the task }
\label{fig:misunderstood}
\end{figure}

We trained a simulated tutor based on this cleaned up data (see below, Section \ref{sec:sim}).

\subsection{Dialogue Actions and their frequencies}
The cleaned up data was annotated for the following dialogue actions:

\begin{itemize}
  \vspace{-0.2cm}
  \item \textbf{Inform}: the action to inform the correct attribute words of an object to the partner, including statement, question-answering, correction, , e.g.\ ``this is a suzuli burchak'' or ``this is sako''; 
  
  \vspace{-0.2cm}
  \item \textbf{Acknowledgment}: the ability to process confirmations from the \textit{tutor}/the \textit{learner}, e.g.\ ``Yes, it's a square".
  
  \vspace{-0.2cm}
  \item \textbf{Rejection}: the ability to process negations from the \textit{tutor}, e.g.\ ``no, it's not red";
  
  \vspace{-0.2cm}
  \item \textbf{Asking}: the action to ask WH or polar questions  requesting correct information, e.g.\ ``what colour is this?'' or ``is this a red square?''.
  
  \vspace{-0.2cm}
  \item \textbf{Focus}: the action to switch the dialogue topic onto specific objects or attributes, e.g.\ ``let's move to shape now";
  
  \vspace{-0.2cm}
  \item \textbf{Clarification}: the action to clarify the categories for particular attribute names, e.g.\ ``this is for color not shape";
  
  \vspace{-0.2cm}
  \item \textbf{Checking}: the action to check whether the partner understood, e.g.\ ``get it?";  
  
  \vspace{-0.2cm}
  \item \textbf{Repetition}: the action to request Repetitions to double-check the learned knowledge, e.g.\ ``can you repeat the color again?"; 
  
  \vspace{-0.2cm}
  \item \textbf{Offer-Help}: the action to help the partner answer questions, occurs frequently when the learner cannot answer it immediately, e.g.\ ``L: it is a ... T: need help? L: yes. T: a sako burchak."; 
\end{itemize}

\begin{figure*}[!ht]
  \subfloat[Dialogue Turns Distribution\label{fig:dialogLength}]
  {\includegraphics[width=0.5\linewidth]{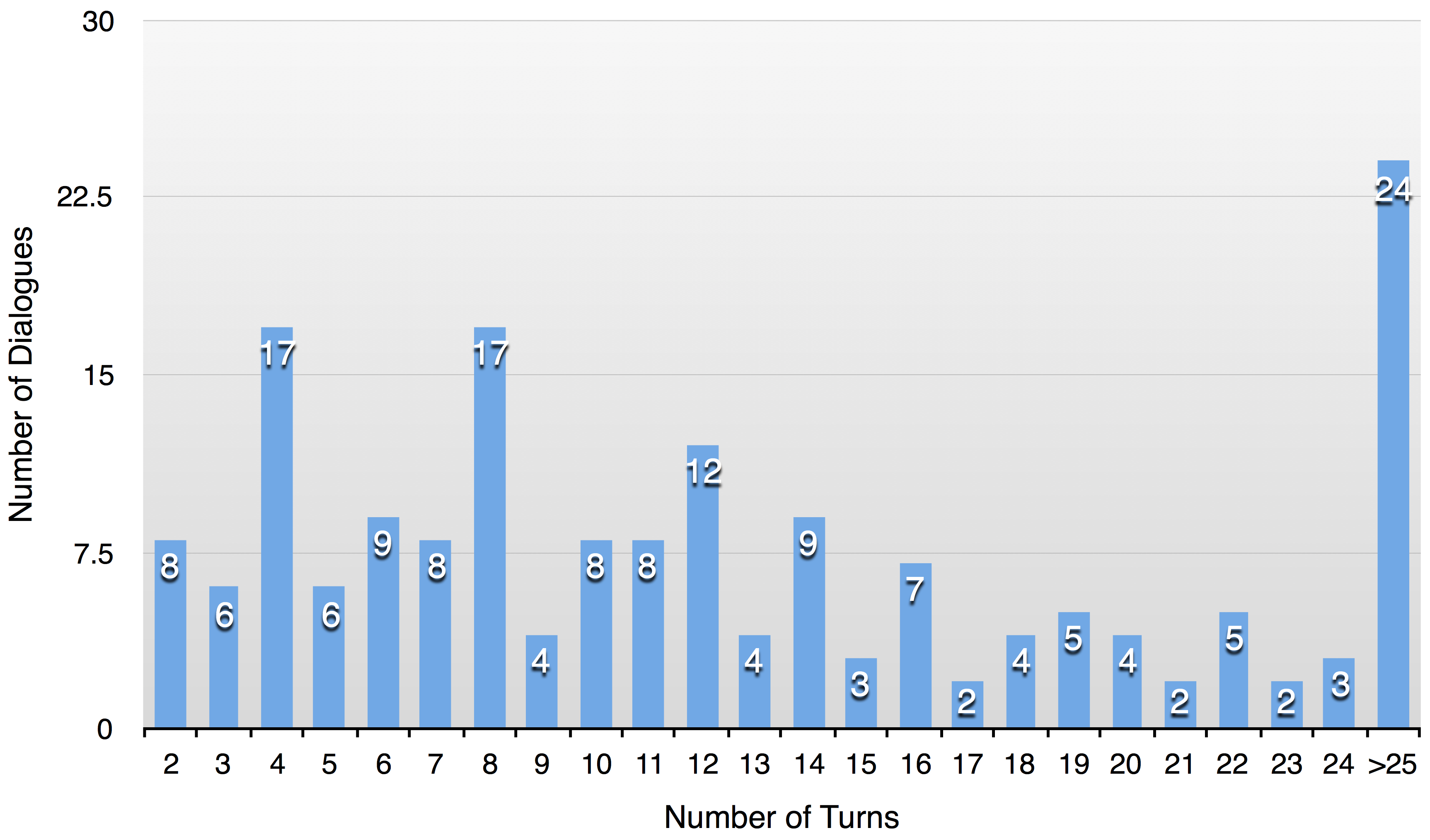}} \hfill
  \subfloat[Dialogue Actions per Turn Distribution\label{fig:actNum}]
  {\includegraphics[width=0.5\linewidth]{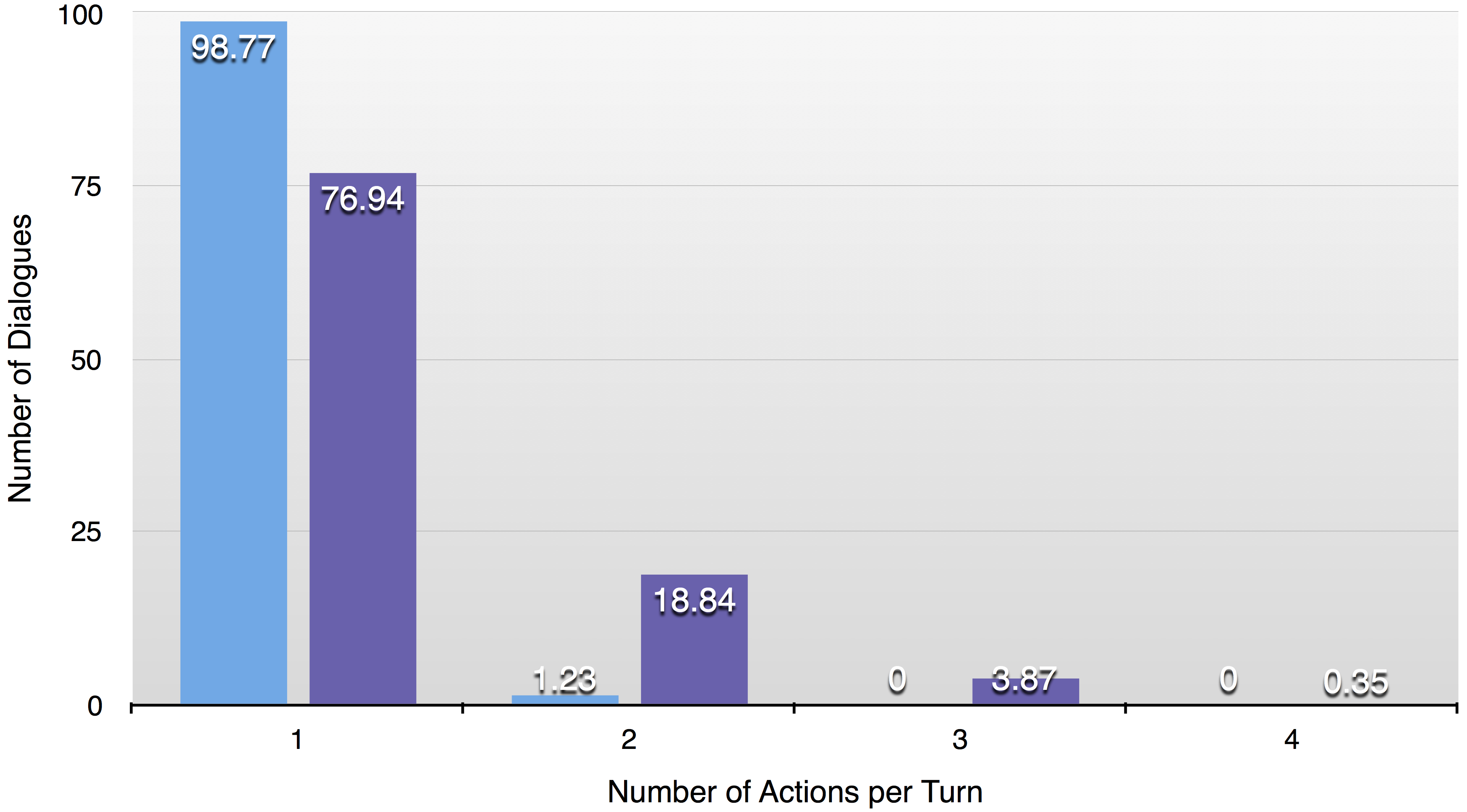}} \hfill

  \subfloat[Dialogue Action Frequencies\label{fig:actFrequency}]
  {\includegraphics[width=0.5\linewidth]{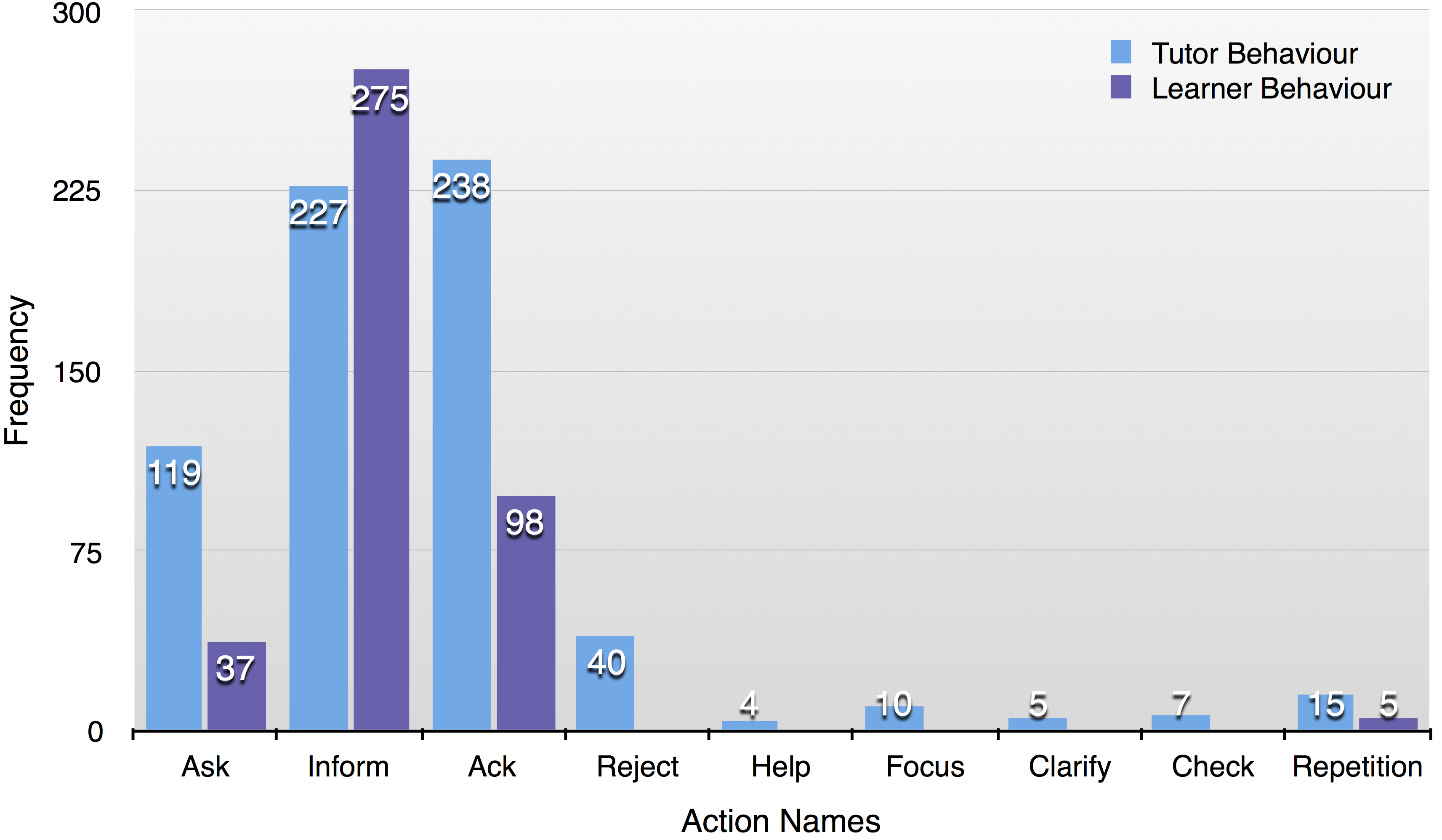}} \hfill
  \subfloat[Incremental Dialogue Phenomena Frequencies\label{fig:phenomenaFrequency}]
  {\includegraphics[width=0.5\linewidth]{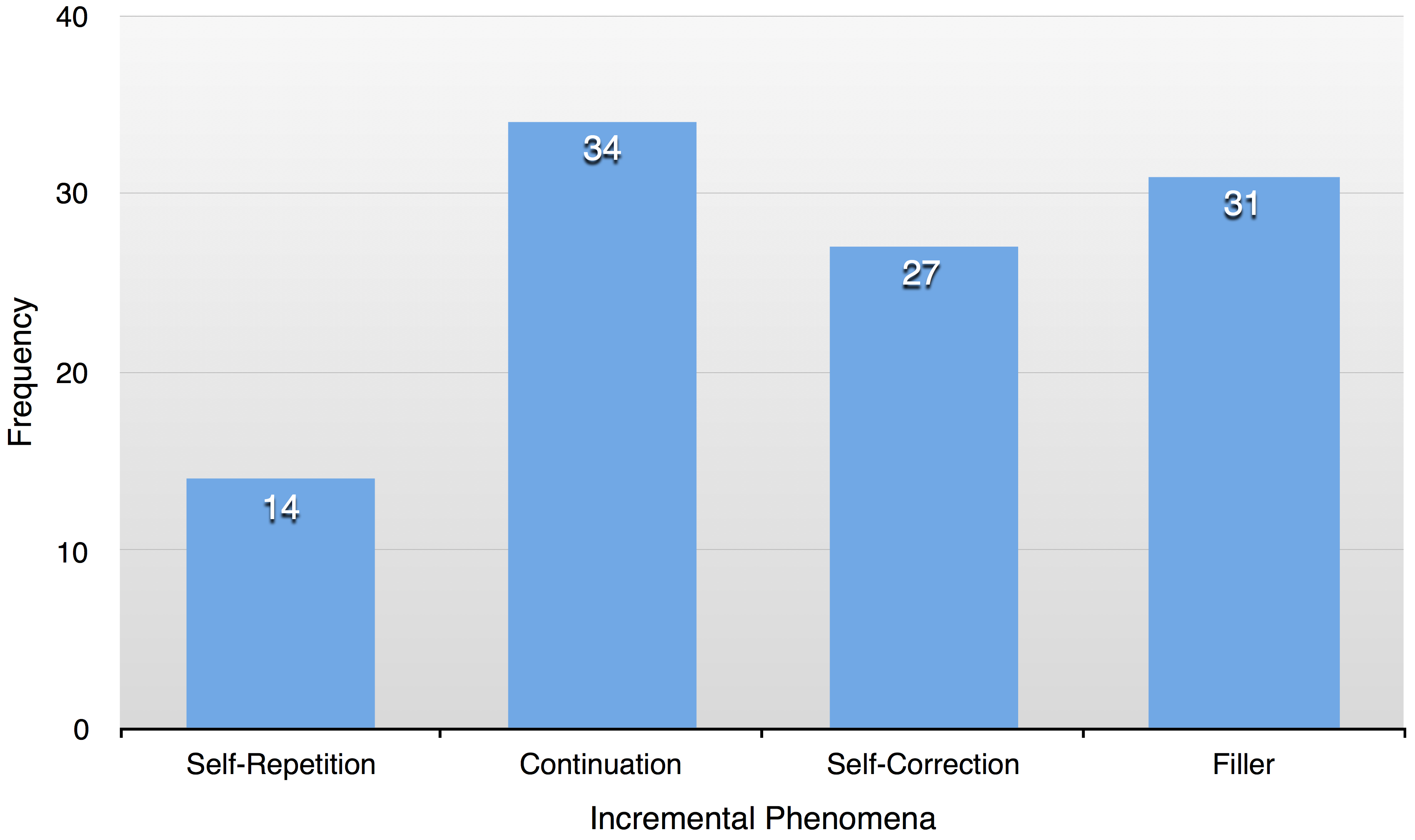}} 

\caption{Corpus Statistics }\label{fig:statistic}
\vspace{-0.2cm}
\end{figure*}

Fig. \ref{fig:actFrequency} shows how often each dialogue action occurs in the data set; and Fig. \ref{fig:actNum} shows the frequencies of these actions by the learner and the tutor individually in each dialogue turn. In contrast with a lot of previous work which assumes a single action per turn, here we get multiple actions per turn (see Table \ref{tab:examples})
In terms of the \textit{Learner} behavior, the learner mostly performs a single action per turn. On the other hand, although the majority of the dialogue turns on the tutor side also have a single action, about $22.59\%$ of the dialogue turns perform more than one action.






\section{TeachBot User Simulation}\label{sec:sim}

Here we describe the generic user simulation framework, based on n-grams, for building user simulation from this type of incremental corpus. We apply this framework to train a TeachBot user simulator that is used to train a RL interactive concept learning agent, both here, and in future work. The model is here trained from the cleaned up version of the corpus. 

\subsection{The N-gram User Simulation}

The proposed user model is a compound n-gram simulation that the probability ($P(t| w_1,..,w_n,c_1,..,c_m)$) of an item $t$ (an action or utterance from the tutor in our work) is predicted based on a sequence of the most recent words ($w_1, \ldots, w_n$) from the previous utterance and additional dialogue context parameters $C$:

\vspace{-0.3cm}
\begin{equation}\label{equ:ngram}
	\small
 	P(t| w_1,..,w_n,c_1,..,c_m) = \frac{freq(t,w_1,..,w_n, c_1,..,c_m)}{freq(w_1,..,w_n, c_1,.., c_m)}
\end{equation}

where $c_1, .., c_m \in C$ represent additional conditions for specific user/task goals (e.g. goal completion as well as previous dialogue context). 

For this specific task, the additional dialogue conditions ($C$) are as follows: (1) the color state ($C_{state}$) for whether the color attribute is identified correctly, (2) the shape state ($S_{state}$) for whether the shape attribute is identified correctly, as well as 3) the previous context ($preContxt$) for which attribute (colour or shape) is currently under discussion.
\vspace{0.1cm}

In order to reduce mismatch risk, the simulation model is able to back-off to smaller n-grams when it cannot find any n-grams matched to the current word sequence and conditions. To eliminate the search restriction by the additional conditions, we applied the nearest neighbors algorithm to search for the n-gram matches by calculating the Hamming distance of each pair of n-grams. 

The n-gram user simulation is generic, as it is designed to handle the item prediction on multiple levels, on which the predicted item, $t$, can be assigned either to (1) a full user utterance ($U_t$) on the utterance level; (2) a combined sequence of dialogue actions ($Das_t$); or alternatively (3) the next word/lexical token. During the simulation, the n-gram model chooses the next item according to the distribution of n-grams. In terms of the action level, a user utterance will be chosen upon a distribution of utterance templates collected from the corpus and combined given dialogue actions $Das_t$. The tutor simulation we train here is at the level of the action and utterance, and is evaluated on the same levels below. However, the framework can be used to train to predict fully incrementally on a word-by-word basis. In this case, the $w_i (i<n)$ in Eq.\ref{equ:ngram} will contain not only a sequence of words from the previous system utterance, but also words from the current speaker (the tutor itself as it is generating).

The probability distribution in equation \ref{equ:ngram} is induced from the corpus using Maximum Likelihood Estimation, where we count how many times each $t$ occurs with any specific combination of the conditions ($w_1,\ldots,w_n,c_1,\ldots,c_m$) and divide this by the total number of times $t$ occurs (see Eq \ref{equ:ngram}).

\subsection{Evaluation of the User Simulation}

We evaluate the proposed user simulation based on the turn-level evaluation metrics by \cite{SimonSpringer2012}, in which evaluation is done on a turn-by-turn basis. Evaluation is done based on the cleaned up corpus (see Section \ref{sec:statistic}). We investigate the performance of the user model on two levels: the utterance level and the action level.

The evaluation is done by comparing the distribution of the predicted actions or utterances with the actual distributions in the data. We report two measures: the Accuracy and Kullback-Leibler Divergence (cross-entropy) to quantify how closely the simulated user responses resemble the real user responses in the BURCHAK corpus. Accuracy ($Acc$) measures the proportion of times an utterance or dialogue act sequence ($Das_t$) is predicted correctly by the simulator, given a particular set of conditions ($w_1,..,w_n, c_1,.., c_m$). To calculate this, all existing combinations in the data of the values of these variables are tried. If the predicted action or utterance occurs in the data for these given conditions, we count the prediction as correct.

Kullback-Leibler Divergence (KLD) ($D_{kl}(P \parallel Q)$) is applied to compare the predicted distributions and  the actual one in the corpus (see Eq.\ref{equ:dkl}). 

\vspace{-0.2cm}
\begin{equation}\label{equ:dkl}
	D_{kl}(P \parallel Q) = \sum\limits_{i=1}^M p_i\log(\frac{p_i}{q_i})
\end{equation}

Table \ref{tab:results} shows the results: the user simulation on both utterance and action levels achieves good performance. The action-based user model, on a more abstract level, would likely be better as it is less sparse, and produces more variation in the resulting utterances. 

Ongoing work involves using BURCHAK to train a word-by-word incremental tutor simulation, capable of generating all the incremental phenomena identified earlier.

\begin{table}[!t]
\centering
\label{my-label}
\begin{tabular}{|c|c|c|}
\hline
\textbf{Simulation} & \textbf{Accuracy (\%)} & \textbf{KLD} \\ \hline
Utterance-level           & 77.98                  & 0.2338                               \\ \hline
Act-level                 & \textbf{84.96}         & \textbf{0.188}                                \\ \hline
\end{tabular}
\caption{Evaluation of The User Simulation on both Utterance and Act levels\label{tab:results}}
\vspace{-0.3cm} 
\end{table}

\vspace{-0.2cm}
\section{Training a prototype concept learning agent from the BURCHAK corpus}
In order to demonstrate how the BURCHAK corpus can be used, we train and evaluate a 
 prototype interactive learning agent using Reinforcement Learning (RL) on the collected data. We follow previous task and experiment settings (see \cite{yu-eshghi-lemon:2016:vl,yu-eshghi-lemon:2016:sigdial}) to compare the learned RL-based agent with a rule-based agent with the best performance from previous work. Instead of using hand-crafted dialogue examples as before, here we train the RL agent in interaction with the user simulation, itself trained from the BURCHAK data as above. 

\vspace{-0.2cm}
\subsection{Experiment Setup} 
To compare the performance of the rule-based system and the trained RL-based system in the interactive learning process, we follow all experiment setup, including visual data-set and cross-validation method. We also follow the evaluation metrics provided by \shortcite{yu-eshghi-lemon:2016:sigdial}
: 
\textit{Overall Performance Ratio} ($R_{perf}$) to measures the trade-offs between the cost to the tutor and the accuracy of the learned meanings, i.e.\ the classifiers that ground our colour and shape concepts. (see Eq.\ref{equ:ratio}).

\vspace{-0.2cm}
\begin{equation} \label{equ:ratio}
 R_{perf} = \frac{\Delta Acc}{C_{tutor}}
\end{equation}

\noindent i.e.\ the increase in accuracy per unit of the cost, or equivalently the gradient of the curve in Fig.\ \ref{fig:result}\ We seek dialogue strategies that maximise this. 

\vspace{0.2cm}
The cost $C_{tutor}$ measure reflects the effort needed by a human tutor in interacting with the system. Skocaj et.\ al.\ \shortcite{Skocaj2009a} point out that a comprehensive teachable system should learn as autonomously as possible,  rather than involving the human tutor too frequently. There are several possible costs that the tutor might incur: $C_{inf}$ refers to the cost (assigned to $5$ points) of the tutor providing information on a single attribute concept (e.g.\ ``this is red'' or ``this is a square"); $C_{ack/rej}$ is the cost ($0.5$ points) for a simple confirmation (like ``yes", ``right") or rejection (such as ``no''); $C_{crt}$ is the cost of correction ($5$ points) for a single concept (e.g.\ ``no, it is blue" or ``no, it is a circle").

\vspace{-0.2cm}
\subsection{Results \& Discussion \label{sec:results}}

  

\ignore{
Fig.\ \ref{fig:learnAccuracy} and \ref{fig:tutorCost} plot the progression of average Accuracy and (cumulative) Tutoring Cost for the rule-based and RL-based systems in the experiment, as the system interacts over time with the Tutor simulation for each of the 500 training instances. As noted in passing, the vertical axes in these graphs are based on averages across the 20 folds - recall that for Accuracy the system was tested, in each fold, at every learning step, i.e.\ after every 10 training instances.
} 

\begin{figure}[!h]
	\centering
 	\includegraphics[width=\linewidth]{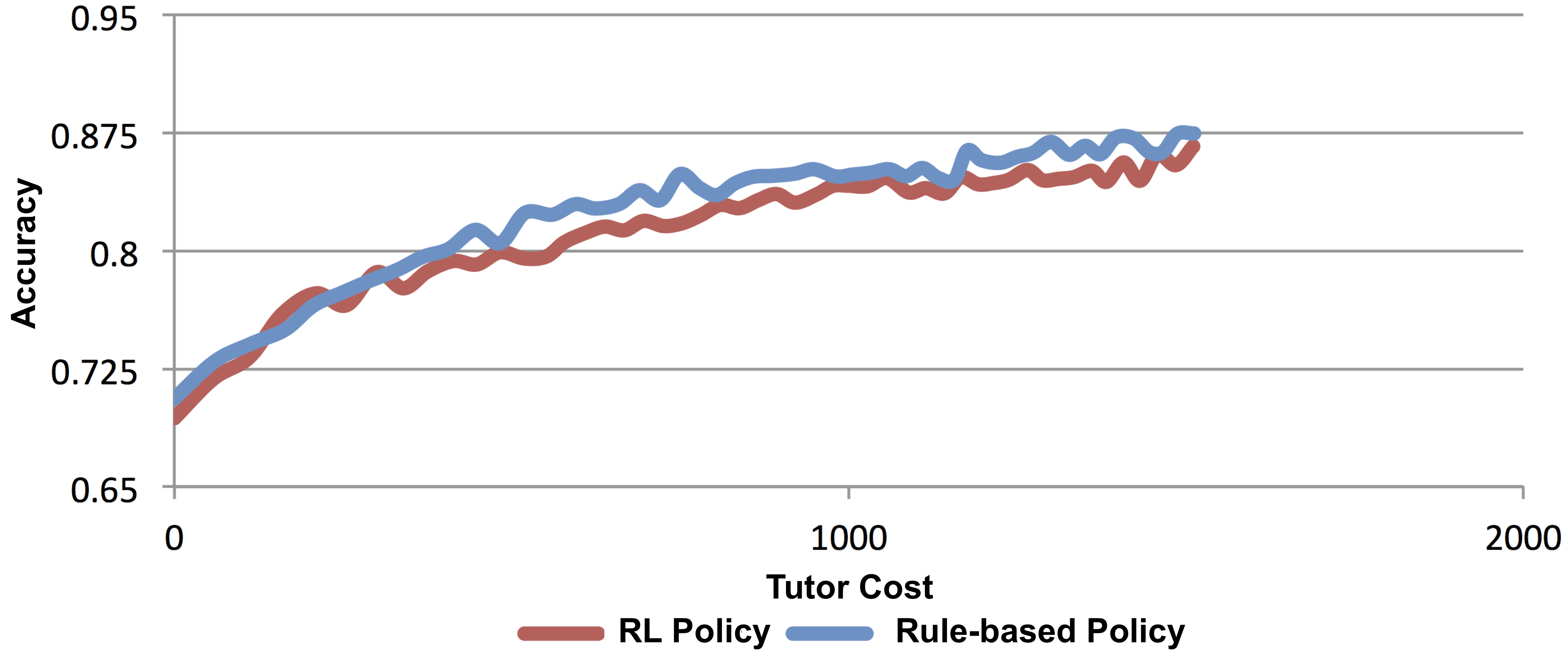}
	\caption{Evolution of Learning Performance}\label{fig:result}
\vspace{-0.2cm}
\end{figure}

Fig.\ \ref{fig:result} plots Accuracy against Tutoring Cost directly. The gradient of this curve corresponds to \textit{increase in Accuracy per unit of the Tutoring Cost}: a measure of the trade-off between accuracy of learned meanings and tutoring cost. 

The result shows that the RL-based learning agent achieves a comparable performance with the rule-based system. 

Table \ref{tab:dlgExample_Simul} shows an example dialogue between the learned concept learning agent and the tutor simulation, where the user model simulates the tutor behaviour (\textit{T}) for the learning tasks. In this example, the utterance produced by the simulation involves two incremental phenomena, i.e. a self-correction and a continuation, though note that these have not been produced on a word-by-word level.

\begin{table}[!ht]
\centering
\begin{tabular}{|l|}
\hline
\begin{tabular}[c]{@{}l@{}}L: so is this shape square? \qquad \qquad \qquad\\ 
						   T: no, it's a squ ... sorry ... a circle. and color? \\
                           L: red? \\ 
                           T: yes, good job.
\end{tabular} \\ \hline    

\end{tabular}
\caption{Dialogue Example between a Learned Policy and the Simulated Tutor \label{tab:dlgExample_Simul}}
\vspace{-0.3cm}
\end{table}

\vspace{-0.2cm}
\section{Conclusion}
We presented a new data collection tool, a new data set,  and and associated dialogue simulation framework which focuses on visual language grounding and natural, incremental dialogue phenomena. The tools and data are freely available and easy to use.

We have collected new  human-human dialogue data on visual attribute learning tasks, which are then used to create a generic n-gram user simulation for   future research and development. We used this n-gram user model to train and evaluate an optimized dialogue policy, which learns grounded word meanings from a human tutor, incrementally, over time.  This dialogue policy optimisation learns a complete dialogue control policy from the data, 
in contrast to  earlier work \cite{yu-eshghi-lemon:2016:sigdial} which only optimised confidence thresholds, and where dialogue control was entirely rule-based.

Ongoing work further uses the data and    simulation framework here to train a word-by-word incremental tutor simulation, with which to learn complete, incremental dialogue policies, i.e.\ policies that choose system output at the lexical level \cite{Eshghi.Lemon.2014}. To deal with uncertainty this system in addition takes all the visual classifiers' confidence levels directly as features in a continuous space MDP.

\vspace{-0.1cm}
\section*{Acknowledgments}
 This research is  supported by the EPSRC, under grant number EP/M01553X/1 (BABBLE project\footnote{\url{https://sites.google.com/site/hwinteractionlab/babble}}),
and by the European Union's Horizon 2020 research and innovation programme under grant agreement No.\ 688147 (MuMMER project\footnote{\url{http://mummer-project.eu/}}).

\bibliographystyle{eacl2017}
\bibliography{bibs/all,bibs/babble,bibs/thesis_all}

\end{document}